**Think it, Run it: Autonomous ML pipeline generation via self-healing multi-agent AI**

Adela BÂRA[1], Gabriela DOBRIȚA[1], Simona-Vasilica OPREA[1,*]
[1]Bucharest University of Economic Studies, Department of Economic Informatics and Cybernetics, no.6 Piața Romană, Bucharest, 010374, Romania, *Corresponding author: simona.oprea@csie.ase.ro

**Abstract:** The purpose of our paper is to develop a unified multi-agent architecture that automates end-to-end machine learning (ML) pipeline generation from datasets and natural-language (NL) goals, improving efficiency, robustness and explainability. A five-agent system is proposed to handle profiling, intent parsing, microservice recommendation, Directed Acyclic Graph (DAG) construction and execution. It integrates code-grounded Retrieval-Augmented Generation (RAG) for microservice understanding, an explainable hybrid recommender combining multiple criteria, a self-healing mechanism using Large Language Model (LLM)-based error interpretation and adaptive learning from execution history. The approach is evaluated on 150 ML tasks across diverse scenarios. The system achieves an 84.7% end-to-end pipeline success rate, outperforming baseline methods. It demonstrates improved robustness through self-healing and reduces workflow development time compared to manual construction. The study introduces a novel integration of code-grounded RAG, explainable recommendation, self-healing execution and adaptive learning within a single architecture, showing that tightly coupled intelligent components can outperform isolated solutions.

**Keywords:** autonomous data science; multi-agent system; code-grounded semantic analysis; ML pipeline automation; self-healing systems; RAG

## 1. Introduction

Modern machine learning (ML) workflows are no longer monolithic programs developed end-to-end, but rather compositions of heterogeneous, reusable components spanning data preprocessing, feature engineering, model training and evaluation. While this modularization enables flexibility and reuse, it also introduces a significant barrier: constructing a functional ML pipeline now requires discovering appropriate components, understanding their behavior and constraints, configuring them correctly and ensuring compatibility across stages. For domain experts without deep software engineering expertise, this process remains time-consuming, error-prone and heavily dependent on manual intervention.

Recent advances in Large Language Models (LLMs) and agent-based systems have enabled natural language (NL) interfaces for automating parts of the data science lifecycle, including pipeline generation and code synthesis. However, existing approaches typically assume that components are either predefined, well-documented or directly generated by the model. In practice, real-world ML ecosystems increasingly rely on user-contributed microservices, where documentation is often incomplete, outdated or inconsistent with the actual implementation. This creates a fundamental challenge: *how can a system reliably discover and compose components when their descriptions cannot be trusted?*

The paper addresses this problem by shifting the source of truth from documentation to implementation. Instead of relying on user-provided descriptions, we propose *a code-grounded approach* in which microservice source code is automatically analyzed to infer semantic capabilities, input-output behavior and usage context. These representations enable robust component discovery even in the absence of reliable documentation and provide a foundation for intelligent pipeline construction. Building on this idea, the paper proposes an *end-to-end autonomous data science framework* that integrates code-grounded semantic analysis, hybrid recommendation, multi-agent orchestration and execution-based learning into a unified system. The framework transforms a dataset and an NL goal into an executable ML pipeline through a sequence of coordinated agents responsible for data understanding, intent formalization, component selection, pipeline construction and execution. To ensure practical robustness, the system incorporates a *self-healing execution mechanism* that recovers from runtime failures by interpreting errors and selecting alternative components, as well as an *adaptive learning module* that continuously improves recommendations based on accumulated execution history.

The main insight of our work is that reliable automation of ML pipelines requires addressing three sources of uncertainty simultaneously: (i) semantic uncertainty, arising from unreliable documentation; (ii) selection uncertainty, arising from multiple plausible component choices; and (iii) execution uncertainty, arising from runtime incompatibilities and failures.

## 2. Literature review

The problem this paper addresses sits at the intersection of four research traditions that have, until now, developed in parallel without fully acknowledging their mutual dependencies. *Software component retrieval* has long grappled with the vocabulary mismatch between how users describe their needs and how developers describe their implementations. *Code representation learning* has progressively moved from syntactic pattern matching toward semantic understanding, culminating in transformer-based models that can reason across programming and NL simultaneously. *Recommender systems research* has studied the cold-start problem, developing hybrid approaches that combine collaborative signals with content-based features, yet consistently assuming that those content features are reliable.

The *vocabulary mismatch* problem, the persistent gap between the terms a searcher uses and the terms a component author chose, has driven software retrieval research for over three decades. (Prieto-Díaz, 1991) introduced faceted classification as a principled solution, organizing software components into controlled vocabulary schemes across multiple independent dimensions of description. (Mili and Mili, 1994) extended this with refinement-based retrieval systems that enabled formal specification matching between queries and component descriptions. Both approaches shared a common assumption: that if authors used standardized vocabulary and precise specifications, retrieval would improve. The assumption proved fragile in practice. Annotation quality depends on author diligence, domain expertise and documentation culture, none of which can be reliably controlled at catalog scale, and neither faceted schemes nor specification languages saw the adoption rates their proponents anticipated.

Neural approaches transformed the field by replacing vocabulary matching with distributed semantic representations. (Gu et al., 2018) introduced Deep Code Search, a dual-encoder architecture that jointly embeds code and NL queries into a shared vector space, enabling retrieval based on semantic proximity rather than lexical overlap. This work established the foundational principle that code and NL, though structurally different, can be projected into a common representational space where meaningful comparison is possible. (Cambronero et al., 2019) extended this line with simpler architectures that achieved comparable accuracy, suggesting that the critical innovation was the joint embedding principle rather than the specific network design.

The CodeSearchNet benchmark (Husain et al., 2020) paired two million functions with their documentation strings to create a training corpus large enough for high-capacity neural models. The benchmark's construction methodology, however, introduced a structural assumption that has gone largely unexamined: that documentation reliably describes the behavior of the code it accompanies. The authors further observed that the training data constructed from code documentation is "not a good match for the code search task", a finding they attribute to vocabulary and style mismatch between documentation and real search queries, but which points equally toward a deeper reliability problem. However, (Robillard and Deline, 2011) provided the behavioral grounding for this concern. Surveying over 440 professional developers at Microsoft, they found that documentation-related obstacles were rated as the most severe barriers to API learning. Developers routinely reported reading source code directly as a more reliable information source than official documentation.

Additionally, (Alon et al., 2019) demonstrated with code2vec that representing code as paths through abstract syntax trees enables models to learn embeddings that capture semantic properties, predicting method names from implementations with surprising accuracy.

The pre-trained transformer models elevated representational quality: CodeBERT (Feng et al., 2020), trained on six programming languages using both masked language modeling and replaced token detection objectives, produced representations that simultaneously capture syntactic code structure and semantic NL meaning. Evaluated on the CodeSearchNet benchmark, CodeBERT achieved state-of-the-art performance on NL code search and code documentation generation, demonstrating bidirectional capability, mapping NL queries to code for search and supporting code-to-text generation under a generation setup. GraphCodeBERT (Guo et al., 2021) extended this by incorporating data flow graphs into pre-training, capturing not just token-level semantics but the semantic relationships between variables that control program behavior. UniXcoder (Guo et al., 2022) further unified code understanding and generation in a single model, while CodeT5 (Wang et al., 2021) introduced identifier-aware representations that leverage the semantic content of variable and function names. Also, prior code summarization work (Hu et al., 2020) addressed single-function description generation; the present approach targets component-level semantic analysis where emergent behaviors not visible at the function level become describable.

The cold-start problem, the failure of collaborative filtering systems to recommend items with no interaction history, was formally characterized by (Schein et al., 2002) at SIGIR 2002. The standard solution is content-based hybrid recommendation: when interaction history is absent, item content features provide similarity signals that enable recommendation from the moment of upload (Lam et al., 2008), (Gantner et al., 2010). In (Burke, 2002) a comprehensive taxonomy of hybrid recommender systems established the theoretical basis for combining collaborative and content-based signals, showing that each compensates for the other's failure modes.

Modern approaches to cold-start have grown substantially more sophisticated. DropoutNet (Volkovs et al., 2017) introduced a neural architecture that explicitly trains for cold-start scenarios by randomly dropping interaction data during training, forcing the model to rely on content features. (Deldjoo et al., 2021) provide a comprehensive contemporary survey of cold-start approaches, cataloguing data augmentation, transfer learning, meta-learning and hybrid strategies across domains. Also, a recent line of work has explored LLM-based zero-shot recommendation for cold-start scenarios, recognizing that generative models can reason about item properties without any interaction history.

(Aghajani et al., 2019) advanced the empirical characterization of documentation failures in a large-scale study mining 878 documentation-related artifacts across GitHub repositories. Their taxonomy identifies five major issue categories: coverage failures where functionality goes undocumented, content failures where documentation is inaccurate or outdated, structure failures where documentation is poorly organized, automation failures and process failures. Their follow-up study (Aghajani et al., 2020) surveyed practitioners about documentation quality perceptions, finding that inaccurate and obsolete documentation was consistently rated as the most critical quality concern across development roles and organization types. (Ratol and Robillard, 2017) studied comment-code inconsistency as a specific manifestation of documentation drift, finding that comments become semantically inconsistent with the code they describe as that code evolves, and that the rate of inconsistency grows monotonically with repository age. (Tan et al., 2007) demonstrated that comment-code inconsistency in production systems causes real bugs when developers rely on comments rather than reading code, establishing that documentation unreliability has operational consequences beyond mere inconvenience. Further, (Treude and Robillard, 2016) took a constructive approach, demonstrating that augmenting API documentation with information automatically extracted from Stack Overflow improved developer task completion rates.

The reviewed literature highlights substantial progress in software component retrieval, code representation learning, recommender systems and LLM-based automation. However, these research streams have largely evolved in isolation, leading to a set of unresolved challenges that become critical when attempting to build autonomous, end-to-end data science systems.

First, existing component discovery approaches, both traditional and neural, implicitly assume that *documentation provides a reliable description of functionality*. As shown in previous empirical studies, this assumption does not hold in practice, where documentation is frequently incomplete, outdated or inconsistent with the underlying implementation. While code representation learning has improved semantic matching, it often relies on code–documentation pairs for training, inheriting the same limitations when documentation quality is poor. As a result, current retrieval systems remain vulnerable to *semantic misalignment between described and actual behavior*. Second, recommender systems research has addressed the cold-start problem through hybrid approaches that combine content-based and collaborative signals. However, these approaches assume that *content features are accurate and stable*, an assumption violated in software ecosystems. Moreover, existing systems rarely integrate *execution-based evidence*, such as observed success rates or workflow patterns, into the recommendation process, limiting their ability to distinguish between components that are semantically similar but differ in practical reliability or compatibility. Third, recent LLM-based agent systems have demonstrated the potential to automate ML workflows from NL. Despite this progress, these systems typically focus on *pipeline generation at the* *abstraction level of code synthesis or predefined components*, without addressing the challenges of *component discovery, configuration and compatibility in dynamic microservice ecosystems*. In particular, they lack mechanisms for *robust failure handling*, often terminating execution when encountering runtime errors rather than adapting the pipeline. Finally, while some systems incorporate feedback loops, there is limited work on *continuous learning from execution history* in the context of ML pipeline construction. The ability to learn from past workflows, capturing which component combinations succeed or fail under specific conditions, remains underexplored, yet is essential for improving performance over time in real-world deployments. To address these gaps, this paper investigates the following research questions:

- **RQ1:** Can a system autonomously generate and execute complete ML pipelines from a dataset and NL goal, without user intervention in component selection, configuration or error handling?
- **RQ2:** Does code-grounded microservice analysis based on direct source code inspection improve component discovery and selection accuracy compared to documentation-based approaches?
- **RQ3:** Does a hybrid recommendation strategy that integrates semantic similarity, data compatibility and execution history outperform semantic-only ranking in terms of pipeline success?
- **RQ4:** Can learning from execution history improve recommendation quality over time, enabling the system to adapt to observed workflow patterns?
- **RQ5:** To what extent can self-healing mechanisms based on LLM-driven error interpretation improve robustness by recovering from runtime failures?

To answer these questions and address the identified research gaps, this paper makes the following *contributions*: (i) Code-grounded microservice representation. We propose a novel approach that analyzes microservice source code to generate semantic descriptions, capabilities and data interface specifications, enabling reliable component discovery independent of documentation quality; (ii) Hybrid recommendation framework with execution awareness. We design an explainable scoring mechanism that combines semantic similarity, keyword matching, data compatibility and execution patterns, improving component selection beyond purely semantic approaches; (iii) Multi-agent architecture for end-to-end pipeline generation. We develop a five-agent system that transforms a dataset and NL goal into a validated and executable ML pipeline through structured intermediate representations; (iii) Self-healing execution mechanism. We introduce a recovery strategy that leverages LLM-based error interpretation to dynamically select alternative components and reconfigure pipeline steps, enabling graceful handling of runtime failures; (iv) Adaptive learning from execution history. We incorporate a pattern learning subsystem that captures both global and user-specific workflow behaviors, enabling continuous improvement in recommendation quality and robustness.

## 3. Methodology

To address the research gaps identified in Section 2 and answer RQ1–RQ5, the methodology proposes an end-to-end autonomous data science framework that integrates code-grounded component understanding, hybrid recommendation, multi-agent orchestration and adaptive learning within a unified architecture. The central objective is to transform a dataset and an NL goal into a validated and executable ML pipeline, without requiring manual intervention in component discovery, configuration or error handling. It requires: understanding the data, interpreting user intent, selecting appropriate components under incomplete or unreliable information and ensuring robust execution in the presence of runtime failures. To this end, the framework is structured around four tightly integrated capabilities:

- Code-grounded semantic analysis, which extracts functional representations of microservices directly from source code, addressing the limitations of documentation-based discovery (RQ2);
- Hybrid component recommendation, which combines semantic similarity, data compatibility and execution history to improve selection accuracy and pipeline success (RQ3, RQ4);
- Multi-agent pipeline orchestration, which decomposes the end-to-end task into specialized reasoning steps, enabling autonomous pipeline construction from NL input (RQ1);
- Self-healing execution mechanisms, which interpret runtime errors and dynamically adapt the pipeline through alternative component selection (RQ5).

These capabilities are implemented through *a five-agent architecture*, supported by a *microservice ecosystem* that manages component lifecycle, semantic indexing, execution tracking and pattern learning. The system operates as a closed feedback loop in which each pipeline execution generates new evidence that improves future recommendations. Figure 1 provides an overview of the system architecture, illustrating the interaction between the microservice management pipeline, the multi-agent orchestration layer and the execution and learning subsystems. The *Presentation Layer* provides user-facing interfaces for uploading, browsing, executing, and building pipelines. The *Application Layer* coordinates three subsystems: Microservice Management (upload, validation and LLM code analysis), Multi-Agent Orchestration (five specialized agents that transform a NL goal into an executed workflow) and Execution & Tracking (with self-healing retry logic). The *Intelligence Layer* houses the system's reasoning capabilities: code-grounded RAG for semantic retrieval, a hybrid recommendation engine that fuses four weighted signals (keyword matching, semantic similarity, data compatibility and learned execution patterns) and a pattern learning system that tracks both global and per-user workflow preferences. The *Persistence Layer* stores structured metadata in MySQL, vector embeddings in ChromaDB, file artifacts on disk and accumulated pattern statistics. The *Execution Layer* runs microservices in resource-constrained Docker containers with security isolation. Three data flows connect the layers end-to-end: microservice upload and indexing, execution and pattern learning, and pipeline request to completed results.

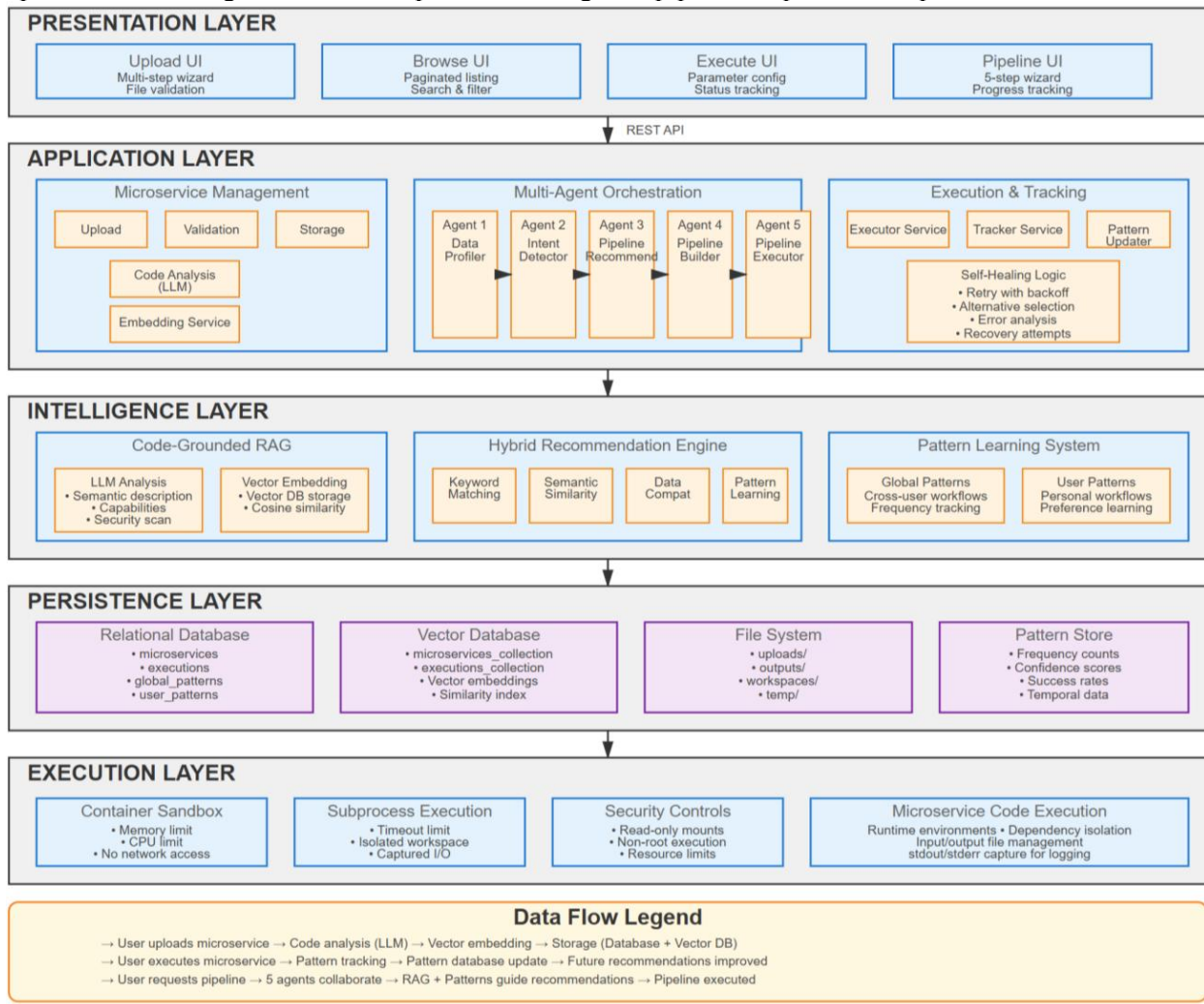

Figure 1. System architecture based on five-layer design for autonomous ML pipeline generation

## 3.1 Microservice ecosystem and code-grounded analysis

The microservice ecosystem manages the complete lifecycle of computational components from initial upload through execution and analysis. The challenge of building an intelligent microservice catalog lies not merely in storing code, but in *understanding* it. The pipeline addresses this challenge through a sequenced four-phase process that transforms a raw code submission into a semantically rich, searchable and securely vetted artifact. The system opens with a multi-step form through which users supply a human-readable name, a functional description, the required Python version, a category classification and semantic keywords. While this metadata serves legitimate purposes, enabling browsing, seeding the recommendation engine and capturing the author's intent, the architecture treats it with appropriate skepticism. User-provided descriptions are frequently incomplete, outdated or inadvertently misleading. Once metadata is collected, the pipeline accepts the actual Python source file and its accompanying requirements specification. Both are staged in a temporary environment, deliberately isolated from the production catalog until they pass a series of validation checks. These checks confirm syntactic validity of the source code, enforce version-pinning in the dependency file to prevent resolution conflicts, impose file size limits against resource exhaustion and perform a surface-level scan for obvious security red flags such as hardcoded credentials or suspicious imports. Rather than relying on what a developer *claims* the code does, the system reads the source directly, typically the first three to ten thousand characters, which generally encompasses the main entry point and central logic and submits it to LLM for structured multi-dimensional analysis. The model produces several interconnected outputs. A two-to-three sentence semantic description captures the microservice's core purpose in terms accessible to non-specialists, while a parallel enumeration of *specific capabilities* provides the granularity that broad claims cannot. The analysis also characterizes input-output data formats, a detail that becomes important when microservices are later composed into pipelines and generates concrete example use cases. A security layer within this phase flags common vulnerability patterns, path traversal risks, SQL injection, command injection and unsafe use of eval or exec, with the explicit caveat that this supplements rather than replaces dedicated program analysis tooling. The results of this analysis are stored alongside the user-provided description in a dual-layer documentation scheme, with the language model version and analysis timestamp recorded for auditability, reproducibility and future re-analysis as models improve. The final phase translates the accumulated textual understanding into a mathematical form suited for similarity search. A composite text representation is constructed from the microservice's name, the AI-generated description (prioritized over the user's), category and keywords, a concatenation that fuses machine-derived understanding with human-chosen terminology. The result is a retrieval system capable of sub-second semantic matching across a catalog of thousands of microservices, that works for *newly uploaded* services with no execution history, solving the cold-start problem that undermines collaborative filtering approaches from the outset.

If the upload pipeline answers the question "*what does this microservice do?*", the execution subsystem answers a deeper one: "*when should it be used, and what comes next?*". By recording every invocation and linking them into chains of purposeful activity, the system transforms a static catalog into something that learns continuously from its own use. At deployment, no patterns exist; recommendations rest entirely on semantic matching. As executions accumulate, chains form. As chains are analyzed, patterns emerge. As patterns strengthen, recommendations improve, steering new users toward workflows that have proven effective for others. Adopted recommendations generate further executions that reinforce successful sequences, while rarely-used or failure-prone transitions are progressively demoted.

**3.2 Multi-agent integration and orchestration**

Code analysis, pattern learning and retrieval-augmented generation are orchestrated into a coherent sequence that takes a user's plain-language intent and returns a fully specified, executable ML pipeline. The specialized agents collaborate in a chain of progressive refinement, each building on its predecessor's output. Figure 2 traces the end-to-end interaction flow of the multi-agent system, illustrating how a user's dataset and NL goal are progressively transformed into executed ML results through the specialized agents.

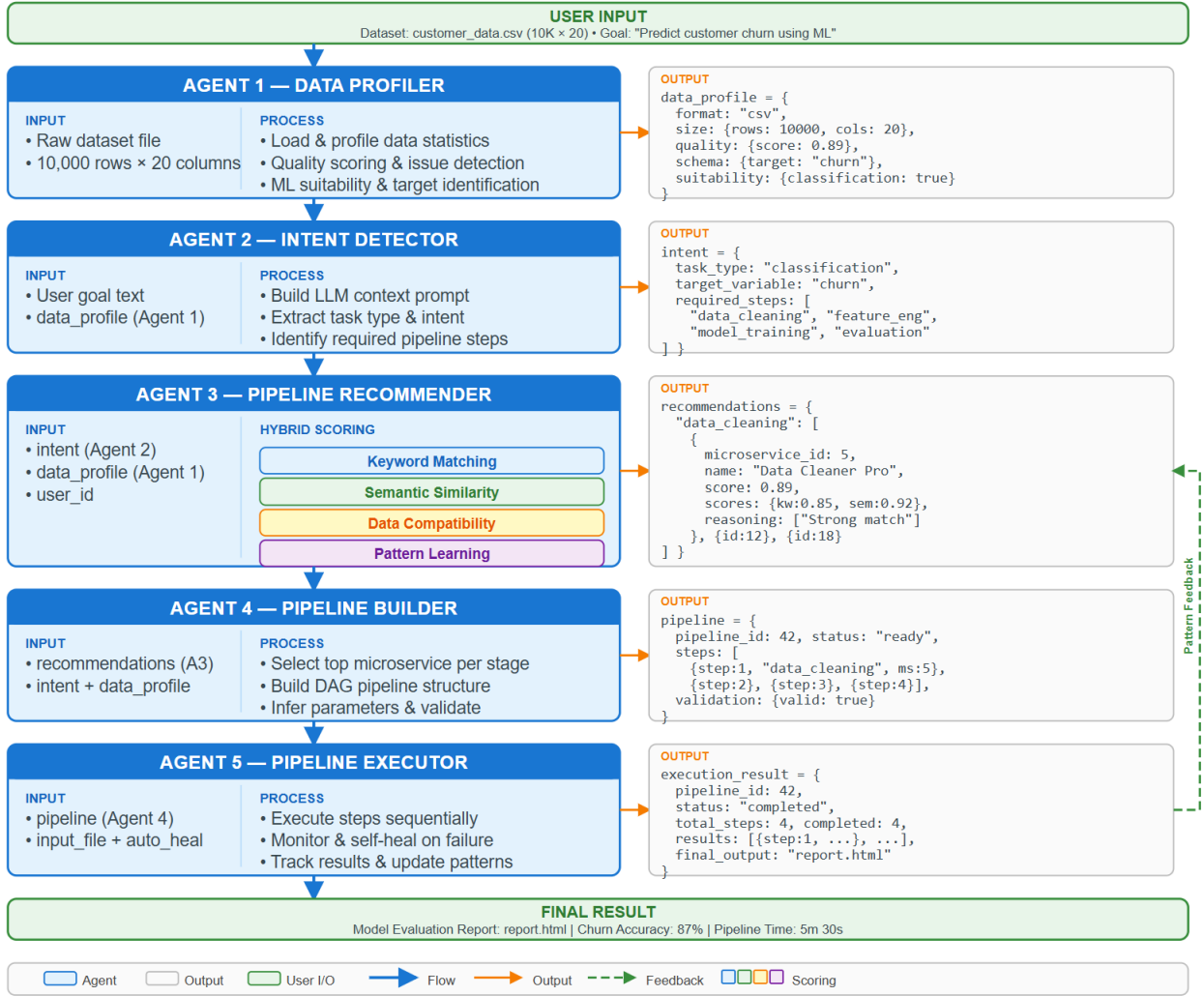

Figure 2. Multi-agent pipeline as a complete interaction flow from user goal to executed ML results

Let $D$ denote the input dataset, $G$ the NL goal provided by the user, $\mathcal{M} = \{m_1, m_2, \ldots, m_n\}$ the set of available microservices and $\mathcal{P}$ a candidate ML pipeline composed of microservices. The objective of the system is to learn a mapping:

$$\mathcal{F}: (D, G) \rightarrow \mathcal{P}^* \tag{1}$$

where $\mathcal{P}^*$ is an executable pipeline that maximizes a utility function:

$$\mathcal{P}^* = \arg\max_{\mathcal{P}} \mathcal{U}(\mathcal{P}, D, G) \tag{2}$$

where $\mathcal{U}(\cdot)$ captures pipeline validity, execution success and output quality. The proposed multi-agent architecture decomposes this mapping into a sequence of structured transformations, each performed by a specialized agent. The system is implemented as a composition of five agents:

$$\mathcal{P} = A_4(A_3(A_2(A_1(D), G))) \tag{3}$$

with execution handled by:

$$R = A_5(\mathcal{P}, D) \tag{4}$$

where $A_1$ is the *Data profiler*, $A_2$-*Intent detector*, $A_3$-*Recommender*, $A_4$-*Pipeline builder* and $A_5$-*Executor*.

*Agent 1-Data profiler: Understanding data*

Before any recommendation can be made, the system must understand the data. The *Data profiler agent* extracts a structured representation of the dataset:

$$A_1: D \rightarrow \Phi \tag{5}$$

where $\Phi$ is a multi-dimensional data profile defined as $\Phi = \{S, T, Q, C, Y\}$ with: S-schema representation (features, types, dimensions); T-statistical properties (distributional metrics); Q-data quality indicators; C-structural characteristics (correlations, redundancy); Y-candidate target variables.

The *Data profiler* ingests the uploaded dataset via pandas, auto-detecting format from file extension and content inspection, then characterizes it across several dimensions simultaneously. Size and schema analysis establishes the structural baseline: row and column counts, file size, pandas-level types and semantic type inference through regex heuristics that detect emails, geographic codes and other domain-specific patterns:

$$S = \{(c_i, type(c_i))\}_{i=1}^{p} \qquad (6)$$

where $c_i$ represents feature (column) $i$, $type(c_i)$ the data type of that column.

For each feature $c_i$, summary statistics are computed:

$$T(c_i) = \{\mu_i, \sigma_i, \min_i, \max_i, \text{cardinality}_i\} \qquad (7)$$

Quality is assessed across three axes: i) completeness (proportion of non-null values); ii) consistency (values within statistically reasonable ranges), iii) uniqueness (duplicate rows, highly correlated columns), combined via a weighted harmonic mean that ensures a single weak dimension meaningfully depresses the overall score rather than being averaged away. Data quality is modeled as a composite score:

$$Q(D) = H(q_{comp}, q_{cons}, q_{uniq}) \qquad (8)$$

where $q_{comp}$-completeness (non-missing ratio); $q_{cons}$-consistency (valid value ranges); $q_{uniq}$-uniqueness (duplicate ratio); $H(\cdot)$-harmonic mean. The correlation structure captures the dependencies between features:

$$C = \{\rho_{ij} = corr(c_i, c_j)\} \qquad (9)$$

Target variable detection applies heuristics over column names, value distributions and temporal patterns to identify likely prediction targets, assigning confidence scores to each candidate. Candidate targets are inferred via heuristic scoring:

$$score(c_i) = \lambda_1 f_{name}(c_i) + \lambda_2 f_{distribution}(c_i) + \lambda_3 f_{temporal}(c_i) \qquad (10)$$

where: $f_{name}$-keyword-based likelihood (e.g., "target", "label"); $f_{distribution}$ - suitability for prediction (binary, continuous); $f_{temporal}$-time-dependence indicator. The most likely target is selected as:

$$c^* = \arg\max_{c_i} score(c_i) \qquad (11)$$

The output is a structured JSON profile consumed by every downstream agent like a shared contract about what the data *is* and what it appears suited for:

$$\Phi = (S, T, Q(D), C, c^*) \qquad (12)$$

which serves as input for downstream agents, particularly intent detection and component recommendation.

The structured profile $\Phi$ reduces uncertainty in subsequent stages by providing a consistent representation of dataset characteristics, enabling more reliable intent inference and component selection.

***Agent 2-Intent detector: Parsing goals into specifications***

With the data profile established, the *Intent detector* receives the user's NL description and uses LLM to translate it into a formal, validated specification. The model is given the full goal statement alongside the data profile and instructed to return structured JSON identifying the ML task type, the target variable, required and optional pipeline stages, constraints around time or interpretability and explicit reasoning for each inference. When a user writes "*predict customer churn*", the model recognizes that *predict* signals supervised learning, that *churn* typically implies binary classification and that the data profile's likely target column is the relevant one. The *Intent detector* transforms the user's goal and the data profile obtained from Agent 1 into a structured task specification:

$$A_2: (G, \Phi) \to I \qquad (13)$$

where $G$ is the NL goal, $I$ is the structured intent representation.

The intent $I$ includes task type $t$, target variable $y$ and the required pipeline stages $S_r$. The task type is inferred as:

$$t = \arg\max_{t' \in \mathcal{T}} P(t' \mid G, \Phi) \qquad (14)$$

where $\mathcal{T}$ is the set of supported task types (e.g., classification, regression, clustering).

For supervised tasks, the target variable is selected as:
$$y = \begin{cases} c^* & \text{if } t \in \{\text{classification, regression}\} \\ \emptyset & \text{otherwise} \end{cases} \quad (15)$$
where $c^*$ is the most likely target identified during data profiling.

The required pipeline stages are derived as:
$$S_r = f(t, \Phi) \quad (16)$$
where $f(\cdot)$ maps the task type and data characteristics to a sequence of processing steps (e.g., data cleaning, feature engineering, model training, evaluation).

The resulting specification then undergoes validation, checking that identified target columns exist, that types match the declared task, that dataset size is sufficient, generating blocking errors or non-blocking warnings accordingly.

*Agent 3-Pipeline recommender: The hybrid scoring engine*

The *Pipeline recommender* is the architectural centerpiece of the multi-agent system. Given the structured intent and data profile, it evaluates every microservice in the catalog against each required pipeline stage:
$$A_3: (I, \Phi, \mathcal{M}) \rightarrow \{\mathcal{M}_s\}_{s \in S_r} \quad (17)$$
where $\mathcal{M} = \{m_1, m_2, \ldots, m_k\}$ is the set of available microservices, $\mathcal{M}_s \subseteq \mathcal{M}$ is the set of candidate microservices for stage $s$.

Each microservice $m \in \mathcal{M}$ is evaluated for a given stage $s$ using a four-signal hybrid scoring algorithm that produces a weighted composite between zero and one:
$$Score(m \mid s) = w_1\, Score_1(m,s) + w_2\, Score_2(m,s) + w_3\, Score_3(m,s,\Phi) + w_4\, Score_4(m,s) \quad (18)$$
where:

- $w_i \in [0,1]$, with $\sum w_i = 1$. In the current implementation, we set the weights as $\{0.3, 0.3, 0.2, 0.2\}$.
- $Score_1(m,s)$ evaluates the keyword-based relevance that operates hierarchically on terms appearing in the microservice *name* score higher than those in the description, on the principle that names signal primary function. Matches are counted and normalized to a zero-to-one scale.
- $Score_2(m,s)$ is the semantic similarity that represents the most important signal. Each microservice is analyzed by LLM at upload time, producing a description of what the code *actually* does. At recommendation time, a query is constructed from the stage name, task type and intent reasoning (for e.g., "*data cleaning classification customer churn prediction with moderate data quality*") embedded into the same space and compared against all microservice vectors via cosine similarity in ChromaDB. Thus, it can be expressed as $Score_2(m,s) = \cos(e_m, e_s)$, where $e_m$ and $e_s$ are embedding representations of the microservice and the stage query.
- $Score_3(m,s,\Phi)$ measures the data compatibility that checks whether the microservice supports the detected data format, whether it addresses the specific quality issues flagged by the profiler (a dataset with quality score below a threshold, for instance, 0.7 favors repair-oriented microservices), and whether required parameters like a target column are present in the data profile.
- $Score_4(m,s)$ expresses the popularity and pattern learning obtained from the execution history. Global patterns (built from every user's chained execution across the platform) are normalized so that 100 successful observed yields a perfect score of 1.0.

For each stage $s$, the top-$k$ candidates are selected as:
$$\mathcal{M}_s = \text{TopK}_{m \in \mathcal{M}}\, Score(m \mid s) \quad (19)$$

The top three candidates per stage are returned with full score breakdowns and human-readable reasoning and not as opaque rankings, but as transparent arguments: "*strong keyword match at 0.85, semantic similarity at 92%, CSV format compatible, 45 successful prior uses*". Thus, the explainability allows users to audit, override and ultimately *trust* the system's judgment.

*Agent 4-Pipeline builder*

The *Pipeline builder* translates the ranked recommendations into a concrete execution plan, resolving input-output compatibility between stages and constructing the DAG of operations. The agent selects the components, arranging them into a dependency graph, inferring the configuration each requires and validating the whole before a single line of code runs. It constructs an executable pipeline from the stage-wise candidate microservices:

$$A_4: \{\mathcal{M}_s\}_{s \in S_r} \rightarrow \mathcal{P} \quad (20)$$

where $\mathcal{P}$ denotes a ML pipeline composed of selected microservices.

The agent can operate in two modes: i) in autonomous mode, it selects the top-ranked candidate for each stage and proceeds without interruption; ii) in interactive mode, it surfaces the ranked options and defers to the user, accommodating domain knowledge or preferences that the scoring algorithm cannot anticipate. Either way, the selected microservices are arranged into a DAG where nodes represent pipeline stages (data cleaning, feature engineering, model training, evaluation, reporting) and edges represent data flow, with each stage's output feeding directly into the next stage's input. Formally, the pipeline is represented as a DAG:

$$\mathcal{P} = (S_r, E) \quad (21)$$

with $S_r$ representing the set of pipeline stages and $E \subseteq S_r \times S_r$ representing data flow dependencies between the stages. Each stage $s \in S_r$ is assigned a selected microservice:

$$m_s^* = \arg\max_{m \in \mathcal{M}_s} Score(m \mid s) \quad (22)$$

Rather than applying uniform defaults or requiring manual specification, the agent constructs a LLM prompt for each pipeline step, providing the microservice's user description and AI-generated code analysis, the stage it fulfills, the ML task type and target variable, the data profile including size and quality scores, and the context of what precedes and follows it in the workflow. The model reasons from this context to infer appropriate parameter values returned as structured JSON ready for direct invocation. For data cleaning steps, quality metrics inform outlier thresholds and missing-value imputation strategies. For feature engineering, task type guides categorical encoding and numeric scaling choices. For model training, dataset size determines whether complex models are viable, and task type constrains algorithm selection. The result is a configured pipeline that reflects the actual characteristics of the data and the goal — not a generic template applied uniformly regardless of context.

For each stage $s$, parameters are inferred as:

$$\theta_s = g(m_s^*, I, \Phi, context_s) \quad (23)$$

where $\theta_s$ are configuration parameters, $context_s$ captures dependencies on previous stages.

This step transforms ranked candidates into a concrete execution plan by selecting optimal components and configuring them based on task intent and data characteristics, resulting in a fully specified pipeline:

$$\mathcal{P} = \{(s, m_s^*)\}_{s \in S_r}, \forall\, m_s^* \in \mathcal{M}_s, \forall s \in S_r \quad (24)$$

Figure 3 details the internal logic of Agent 4, which transforms ranked microservice recommendations into a validated, executable pipeline. The configured pipeline then undergoes a four-part validation (checking stage completeness, data flow compatibility, column references and dataset size) before being persisted to MySQL as a complete pipeline specification ready for Agent 5.

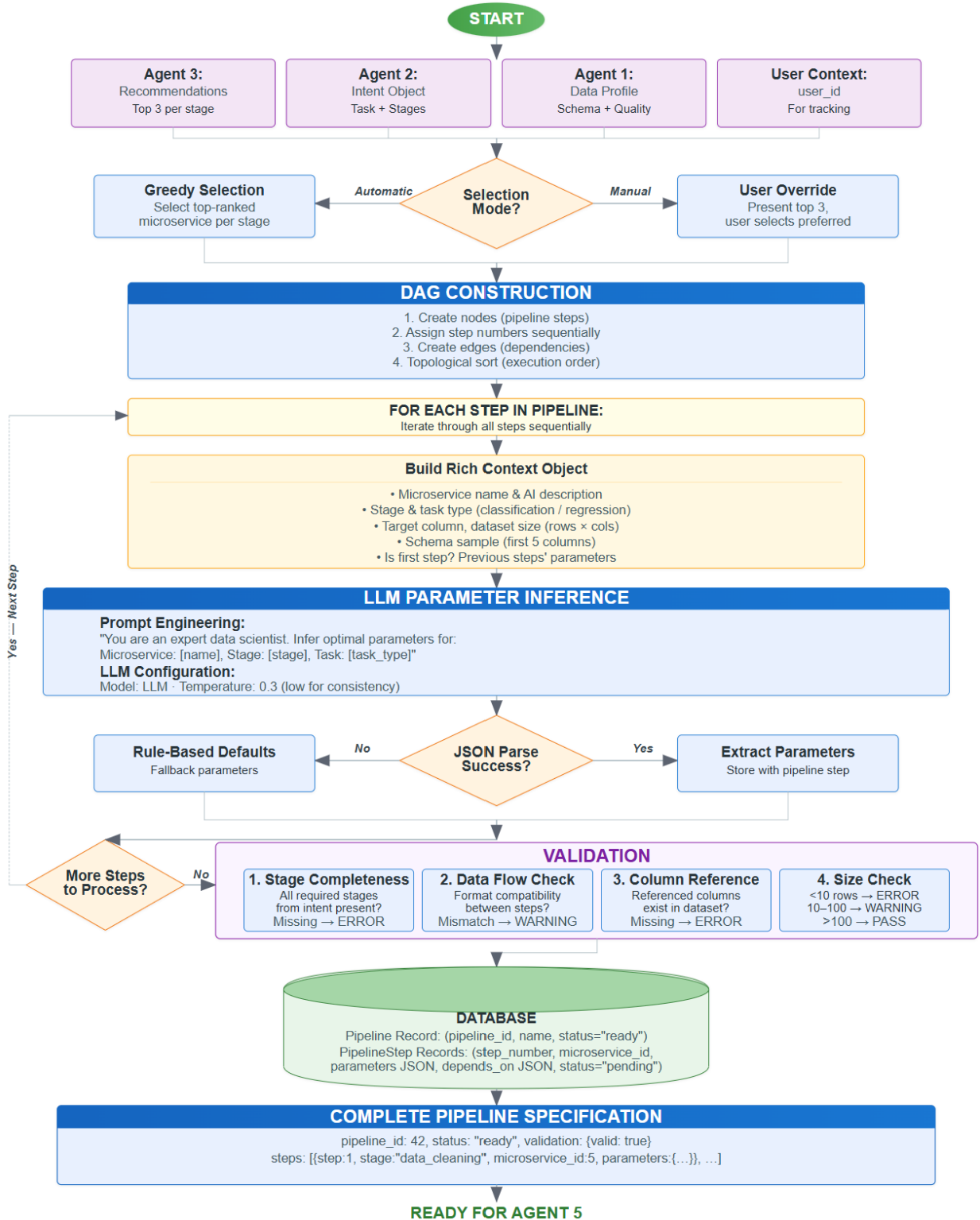

Figure 3. Agent 4-*Pipeline builder*: From ranked recommendations to validated execution plan

***Agent 5-Pipeline executor***

The *Pipeline executor* runs DAG, invoking microservices in sequence within isolated Docker containers, capturing outputs and passing them forward as inputs to subsequent stages. The agent executes the constructed pipeline on the dataset:

$$A_5: (\mathcal{P}, D) \to R \tag{25}$$

where $R$ represents the final outputs (predictions, metrics or analysis results).

Execution proceeds sequentially along the DAG structure:

$$D_{s+1} = m_s^*(D_s, \theta_s) \tag{26}$$

where $D_s$ is the intermediate data at stage $s$; $D_1$ is the initial dataset $D$.

Before the pipeline is marked execution-ready, a validation pass performs several targeted checks. It confirms that every stage identified by the *Intent detector* is present, with no gaps that would break the workflow. It inspects data format compatibility between connected steps, flagging mismatches between expected outputs and required inputs before they become runtime errors. It verifies that the target column exists in the dataset and remains intact through each transformation. It enforces minimum data size requirements, generating hard errors for datasets too small to be meaningful and softer warnings for borderline cases.

The pipelines that pass validation are persisted as a *Pipeline* entity in the database, with associated *PipelineStep* records capturing sequential order, stage assignment, selected microservice, inferred parameters, inter-step dependencies, expected I/O formats and individual step status. The pipeline itself carries a human-readable name derived from the user's original goal, ownership metadata for access control and a status that will progress through *draft*, *ready*, *running*, *completed* and *failed* as execution unfolds. The *Pipeline builder* agent is, in a sense, where abstraction ends and execution begins. It takes the structured output of upstream reasoning (intent, data understanding, ranked recommendations) and produces something concrete: a validated, configured, persisted workflow ready to run. Its parameter inference capability is particularly consequential, as it means users do not need to understand the internal requirements of each microservice they deploy. The system configures itself, using the same contextual reasoning that drove component selection to determine how each component should behave once selected.

Each execution begins by creating an isolated workspace directory, named using the pipeline identifier and a timestamp to prevent collisions. The input dataset is copied into this workspace as the initial artifact, and every subsequent step reads and writes only within that directory, keeping executions reproducible and contained, preventing accidental cross-run contamination and avoiding reliance on shared filesystem state. The agent then loads the ordered *PipelineStep* records from the database and executes them sequentially. Before each invocation, the step is marked as running and the state transition is committed so that progress is visible to users and monitoring services. The executor calls the microservice runtime with fully resolved inputs: the microservice code reference, the current input artifact (the original dataset for step one, otherwise the previous step's output), the output path for the step and the stage parameters inferred by the *Pipeline builder*. Microservices run inside isolated containers (or a subprocess-based runner in constrained deployments), with timeouts and *stdout/stderr* capture to support monitoring and debugging.

When retries fail, the executor enters self-healing mode instead of terminating the pipeline. It asks the recommender for alternative microservices for the same stage, ranked by the same hybrid score, excluding any already tried:

$$m_s^{new} = \arg\max_{m \in \mathcal{M}_s \setminus \mathcal{T}_s} Score(m \mid s) \tag{27}$$

where $\mathcal{T}_s$ is the set of previously attempted microservices for stage $s$.

The executor substitutes the best remaining candidate and reruns the step on the same input artifact; if it succeeds, execution proceeds and the run record notes the substitute. The pipeline is then updated:

$$m_s^* \leftarrow m_s^{new} \tag{28}$$

and execution resumes from stage $s$.

Only after all viable alternatives are exhausted is the pipeline marked failed, with logs and diagnostics preserved, turning the recommender's ranked list into a recovery queue as well as a selection tool.

Execution also closes the system's learning loop. Each run produces structured traces (stage sequence, chosen components, parameters, success/failure outcomes and timing) which are written back to the execution history store. Successful sequences strengthen the corresponding workflow patterns, while failure cases reduce their influence:

$$freq(m, s) \leftarrow freq(m, s) + 1 \tag{29}$$

which influences future recommendations through $Score_4(m,s)$.

The execution process produces results and also enables adaptive learning and robustness through dynamic recovery and continuous feedback integration.

The multi-agent pipeline embodies a self-reinforcing feedback loop. Early in deployment, recommendations lean on keywords and semantic similarity. As executions accumulate, pattern scores grow in influence: microservices that consistently appear in successful pipelines rise in ranking while underperforming ones are naturally demoted. Figure 4 illustrates the three principal data flows that underpin the system. The *Upload and Indexing* flow transforms submitted source code into searchable semantic representations through LLM analysis and vector embedding. The *Execution and Pattern Learning* flow records microservice invocations and automatically chains them into usage patterns, allowing the system to improve its recommendations over time. The *Hybrid Recommendation* flow shows how Agent 3 selects candidate microservices by combining four weighted signals, keyword matching, semantic similarity, data compatibility, and learned execution patterns, in a scoring engine that returns the top-ranked candidates per pipeline stage.

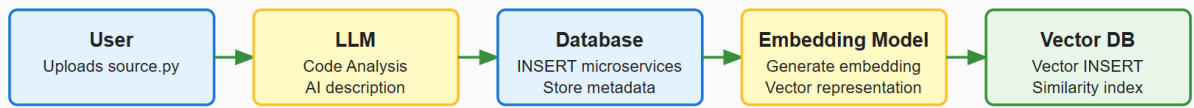

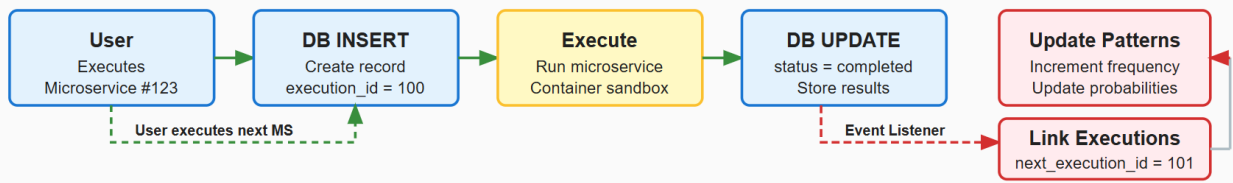

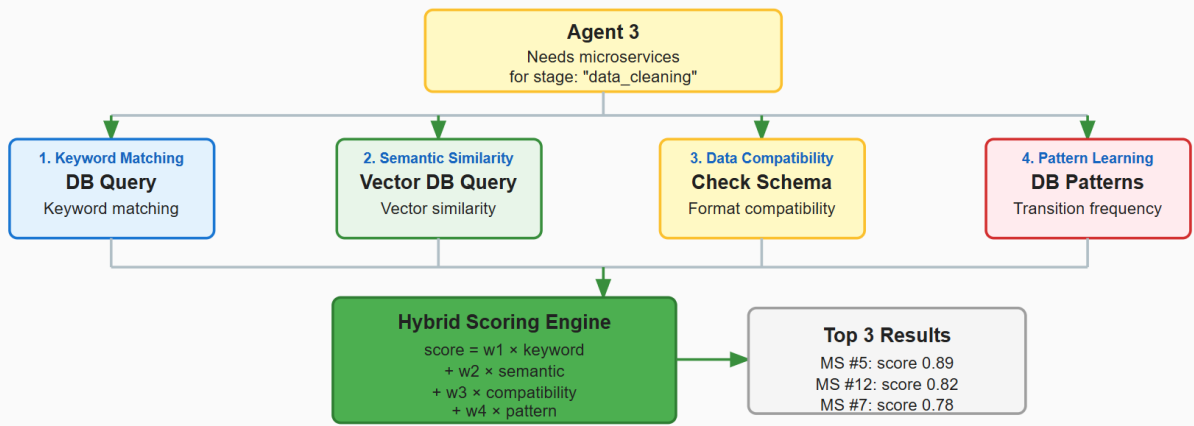

Figure 4. Data flow architecture across: Upload, Execution and Recommendation flows

## 4. Simulations

We evaluate the proposed framework with respect to its core capability: transforming a dataset and an NL goal into executed ML results through automatic discovery and composition of user-uploaded microservices. The evaluation focuses on whether code-grounded component analysis, hybrid recommendation and execution history learning enable reliable end-to-end pipeline generation without manual intervention. The evaluation is structured around the defined RQs, focusing on the system's ability to autonomously generate, execute and improve ML pipelines under realistic conditions.

The unit of analysis is a complete automation attempt, defined as the process from dataset upload and NL goal specification to final ML results (predictions, model metrics or analytical outputs). Each trial simulates realistic usage, where a domain expert provides a dataset and a high-level objective, and the system is required to autonomously produce valid results. We evaluate both operational success (pipeline completion) and output quality (correctness of results).

**4.1 Evaluation metrics**

Three primary metrics are used:
1. Pipeline success rate measures the proportion of dataset–goal pairs for which the system generates and executes a complete pipeline producing valid outputs without manual intervention. This is the primary utility metric.
2. Time-to-results measures wall-clock time from goal submission to final output, reported as median and 90$^{th}$ percentile, capturing both system latency and user wait time.
3. Component selection accuracy evaluates the proportion of correctly selected microservices per pipeline stage, evaluated against expert-constructed reference pipelines.

For supervised tasks, output quality is measured using standard metrics (AUC-ROC, F1 for classification; RMSE, MAE for regression). For unsupervised tasks, clustering quality is evaluated using ARI, NMI and silhouette score. For exploratory analysis tasks, correctness is verified by comparing computed statistics (e.g., correlations, distributions, outlier counts) against reference implementations within a numerical tolerance $\varepsilon = 10^{-6}$.

**4.2 Benchmark and experimental setup**

The experimental implementation instantiates the LLM using GPT-4 for code-grounded analysis, intent interpretation and self-healing reasoning. The model is accessed via API with deterministic settings (temperature = 0.3) to ensure reproducibility across runs. For semantic representation, microservices and stage queries are encoded using *Sentence-Transformers all-MiniLM-L6-v2*, which produces 384-dimensional embeddings optimized for efficient similarity search. Embeddings are stored and queried using ChromaDB, enabling sub-second retrieval over the microservice catalog. These settings reflect a trade-off between semantic expressiveness and computational efficiency, while remaining consistent with widely adopted practices in retrieval-augmented systems.

We construct a benchmark of 150 ML tasks reflecting realistic data science scenarios. Tasks are grouped into three categories:
- Supervised learning (72 tasks): 38 classification and 34 regression problems using OpenML datasets (5K–500K rows, 10–200 features) across domains such as finance, healthcare, e-commerce and social science.
- Unsupervised analysis (48 tasks): clustering, dimensionality reduction and anomaly detection.
- Exploratory data analysis (30 tasks): data profiling, correlation analysis and outlier detection.

Each task consists of a dataset paired with an NL goal (e.g., "*predict customer churn*" or "*identify transaction clusters*"), simulating how domain experts express objectives in practice.

The evaluation uses a catalog of 127 user-uploaded Python microservices representing typical functionality in ML workflows: i) 45 preprocessing components (data cleaning, encoding, scaling, validation); ii) 38 modeling components (classical ML, neural networks, ensembles); iii) 22 evaluation components (metrics, validation strategies); iv) 12 visualization components; v) 10 utility components (data loading, exporting, notifications).

Each microservice includes source code, metadata and user-provided documentation. Documentation quality varies realistically: approximately 30% of components contain incomplete, outdated or misleading descriptions, reflecting real-world conditions where documentation maintenance lags behind implementation.

To evaluate cold-start performance, 23 microservices (18%) are designated as having no prior execution history at the start of evaluation.

To isolate architectural contributions, we evaluate four configurations (Table 1):
A. *Full system* represents our complete architecture: LLM analyzes uploaded microservice source code to generate semantic descriptions, the hybrid scoring function combines semantic similarity with data compatibility and execution patterns, and the system learns continuously from workflow executions.

B. *Ablation-NoHistory* removes execution history learning to isolate its contribution, testing whether the system can operate effectively in cold-start scenarios or whether learning from usage is essential.
C. *Baseline-DocBased* replaces code-grounded analysis with documentation-based discovery, using user-provided README descriptions and docstrings instead of analyzing source code, testing our central hypothesis that code analysis provides more reliable semantic signals than potentially stale documentation.
D. *Baseline-SemanticOnly* removes data compatibility checking and execution patterns, relying solely on semantic similarity for component ranking, testing whether the hybrid scoring function provides measurable value beyond pure semantic matching.

This setting tests whether code-grounded analysis enables immediate discovery and effective use of newly uploaded components without reliance on historical usage patterns.

Table 1. System configurations for ablation analysis

| Configuration | Component Analysis | Scoring Function | Execution History |
|---|---|---|---|
| Full System | Code-grounded (GPT-4) | Hybrid (4 signals) | Enabled |
| Ablation-NoHistory | Code-grounded (GPT-4) | Hybrid (4 signals) | Disabled |
| Baseline-DocBased | Documentation-only | Hybrid (4 signals) | Enabled |
| Baseline-SemanticOnly | Code-grounded (GPT-4) | Semantic similarity only | Disabled |

*Component Analysis* indicates whether microservice semantic descriptions are generated from source code analysis (code-grounded) or extracted from user-provided documentation. *Scoring Function* specifies whether component ranking uses all four signals (keyword matching 30%, semantic similarity 30%, data compatibility 20%, execution patterns 20%) or semantic similarity alone. *Execution History* indicates whether the system learns from accumulated workflow executions.

### 4.3 Evaluation protocols

We design targeted experiments to isolate the contribution of each system component.

### 4.3.1 Cold-start evaluation

We evaluate cold-start capability by measuring pipeline success rate separately for tasks requiring at least one zero-history microservice versus tasks using only components with established execution history. This tests whether code-grounded analysis enables immediate effective use of newly uploaded components, a critical requirement for extensible catalogs. We stratify results by component age (newly uploaded vs. established) and measure the success rate gap. A small gap (within 10 percentage points) indicates effective cold-start handling; a large gap indicates over-reliance on execution history at the expense of new component discovery.

### 4.3.2 Documentation quality sensitivity

We explicitly manipulate documentation quality to test robustness. For the 127 microservices, we create degraded documentation variants at three severity levels: *mild degradation* (remove usage examples and caveats, ~30% information loss), *moderate degradation* (remove parameter descriptions, ~60% information loss) and *severe degradation* (retain only component name and category, ~90% information loss). We rerun the benchmark at each degradation level using both code-grounded and documentation-based discovery, measuring pipeline success rate degradation. Steep degradation curves for documentation-based discovery indicate vulnerability to documentation quality; flat curves for code-grounded discovery indicate robustness, validating that analyzing source code circumvents documentation reliability issues.

### 4.3.3 Learning dynamics over time

We evaluate execution history learning using temporal protocol: tasks at time $t$ may use only execution traces observed before $t$. We partition the 150 tasks into five temporal cohorts of 30 tasks each, executed sequentially with accumulating history. We measure pipeline success rate per cohort, yielding learning curves that demonstrate whether recommendation quality improves as evidence accumulates. We additionally analyze *user-specific learning* by comparing personalized pattern weights (heavily weighting a user's own workflow history) versus global patterns (aggregated across all users), testing whether individual usage patterns provide signal beyond population-level statistics. Upward-sloping learning curves indicate actionable learning from execution history; flat curves indicate that additional evidence provides no benefit. The gap between personalized and global patterns quantifies the value of user-specific adaptation.

### 4.3.4 Self-healing effectiveness

We inject realistic failure conditions into 20% of pipeline executions to test recovery mechanisms: type mismatches (component expects array, receives DataFrame), missing required parameters (component needs target column name, not provided), numerical instability (overflow, NaN propagation) and resource exhaustion (out of memory, timeout). We compare two recovery strategies: *retry-only* (repeat failed stage up to 3 times with exponential backoff, then abort) and *LLM-based self-healing* (use GPT-4 to interpret error message, reason about root cause and select alternative microservice from ranked candidates with parameter adaptation). We measure recovery rate (proportion of injected failures successfully resolved), added latency (extra time required for recovery) and net impact on overall pipeline success rate. High recovery rate with acceptable latency demonstrates that intelligent alternative selection enables graceful degradation rather than catastrophic failure, improving system robustness for production deployment.

### 4.4 Results

We present results organized by research question (as in Table 2), focusing on the system's core capability: autonomous ML pipeline generation from NL goals. Statistical significance is assessed using paired t-tests ($\alpha=0.05$) across 150 task executions, comparing system configurations and measuring improvement over baselines.

Table 2. Evaluation protocol summary mapping research questions to experimental protocols and metrics

| RQ | Evaluation Protocol | Primary Metrics |
|---|---|---|
| RQ1: End-to-End Automation | 150 tasks (dataset+NL goal→results) | Pipeline success rate; time-to-results; output quality (AUC, RMSE) |
| RQ2: Code-Grounded Discovery | Full system vs. Baseline-DocBased | Success rate gap; component selection accuracy |
| RQ3: Hybrid Scoring | Full system vs. Baseline-SemanticOnly | Success rate improvement; ablation analysis |
| RQ4: Execution History | Temporal evaluation (5 cohorts) | Learning curve slope; success rate improvement over time |
| RQ5: Self-Healing | Failure injection (20% of runs) | Recovery rate; added latency; net success rate impact |

#### 4.4.1 RQ1: End-to-end autonomous pipeline generation

Table 3 presents aggregate performance across 150 ML tasks. The full system achieves 84.7% pipeline success rate, meaning that for 127 of 150 dataset-goal pairs, the system autonomously discovered appropriate microservices, constructed valid pipelines, executed them successfully and produced ML results, without user intervention in component selection, parameter configuration or error handling.

Table 3. End-to-end pipeline generation performance across 150 tasks

| Configuration | Success Rate | Median Time (s) | P90 Time (s) | Avg. Output Quality |
|---|---|---|---|---|
| Full System | 84.7% (127/150) | 143 | 487 | AUC: 0.81, RMSE: 0.24 |
| Ablation-NoHistory | 79.3% (119/150) | 156 | 521 | AUC: 0.79, RMSE: 0.26 |
| Baseline-DocBased | 68.0% (102/150) | 189 | 673 | AUC: 0.76, RMSE: 0.29 |
| Baseline-SemanticOnly | 71.3% (107/150) | 167 | 589 | AUC: 0.77, RMSE: 0.27 |
| Manual Construction | 98.0% (147/150) | 3,240 | 5,890 | AUC: 0.84, RMSE: 0.21 |

Compared to baselines, the full system shows consistent improvements. Removing execution history reduces success to 79.3%, indicating that learned workflow patterns contribute to robustness. Documentation-based discovery performs substantially worse (68.0%), confirming that reliance on user-provided descriptions leads to degraded component selection. Semantic-only scoring achieves 71.3%, demonstrating that semantic similarity alone is insufficient for reliable pipeline construction. Manual pipeline construction achieves higher success (98.0%) but requires significantly more time (median 3,240 seconds vs. 143 seconds), corresponding to a 22.7× speedup for the proposed system. Output quality remains competitive (AUC 0.81 vs. 0.84; RMSE 0.24 vs. 0.21).

Performance varies across task types (Table 4), with classification achieving the highest success rate (89.5%) due to well-established workflow patterns, while dimensionality reduction shows slightly lower performance (80.0%) due to higher sensitivity to data representation. The remaining failures (15.3%) are primarily due to ambiguous NL goals, gaps in microservice catalog coverage, runtime incompatibilities not detectable through static analysis. The results indicate that limitations arise from problem specification and component availability rather than pipeline construction itself.

Table 4. Success rates stratified by ML task type

| Task Category | Count | Full System | Doc-Based | Example Failed Goal |
|---|---|---|---|---|
| Classification | 38 | 89.5% (34/38) | 71.1% (27/38) | "Predict customer churn"→selected incompatible encoder |
| Regression | 34 | 85.3% (29/34) | 67.6% (23/34) | "Forecast sales"→missing time-series preprocessing |
| Clustering | 28 | 82.1% (23/28) | 67.9% (19/28) | "Segment customers"→wrong distance metric |
| Dimensionality Reduction | 20 | 80.0% (16/20) | 65.0% (13/20) | "Visualize embeddings"→incompatible input shape |
| EDA | 30 | 83.3% (25/30) | 66.7% (20/30) | "Find anomalies"→statistical test mismatch |

#### 4.4.2 RQ2: Code-grounded component discovery

We manually evaluate whether selected microservices are appropriate for each pipeline stage, comparing automatic selections against gold-standard pipelines constructed by ML experts. Table 5 presents accuracy by pipeline stage.

Table 5. Component selection accuracy by pipeline stage

| Pipeline Stage | Full System (Code) | Baseline (Docs) | Δ Accuracy | Typical Error (Docs) |
|---|---|---|---|---|
| Preprocessing | 91.3% | 72.7% | +18.6 pp | Stale docs: removed parameters not mentioned |
| Feature Engineering | 88.7% | 69.3% | +19.4 pp | Incomplete docs: edge cases undocumented |
| Modeling | 89.3% | 74.0% | +15.3 pp | Misleading docs: hyperparameters misrepresented |
| Evaluation | 94.0% | 81.3% | +12.7 pp | Outdated docs: metric definitions changed |
| Overall Pipeline | 87.3% | 70.7% | +16.6 pp | Accumulates across stages |

Feature engineering shows the largest accuracy gap (+19.4 pp), reflecting that custom domain-specific transformations are particularly prone to documentation drift-developers modify transformation logic but forget to update READMEs. Evaluation metrics show the smallest gap (+12.7 pp) because evaluation microservices tend to have simpler, more stable interfaces that are easier to document accurately.

Code-grounded analysis provides substantially more reliable component discovery than documentation-based approaches. By using LLM to read and analyze uploaded Python source code, the system generates semantic descriptions that reflect actual implementation behavior rather than potentially stale author-written documentation. The 16.7 pp success rate advantage and 16.6 pp selection accuracy improvement demonstrate that documentation unreliability (a well-documented phenomenon in software engineering) has measurable, substantial operational impact on ML pipeline automation systems, validating the architectural decision to treat source code, not documentation, as the authoritative signal for component capabilities.

#### 4.4.3 RQ3: Hybrid scoring vs. semantic-only ranking

The hybrid scoring approach improves pipeline success from 71.3% to 84.7% (+13.4 pp, $p < 0.001$). Failure analysis (Table 6) shows that hybrid scoring prevents 76% of type mismatches, 82% of missing data errors and 67% of format incompatibilities.

Table 6. Pipeline failure analysis comparing semantic-only ranking against full hybrid scoring

| Failure Type | Semantic-Only Failures | Full System Failures | Example |
|---|---|---|---|
| Type/Shape Mismatch | 18 | 4 | Scaler expects 2D array, component outputs 1D |
| Missing Required Data | 11 | 2 | Model needs target column, not in dataset |
| Format Incompatibility | 9 | 3 | Component expects CSV, receives DataFrame |
| Parameter Conflicts | 5 | 1 | Normalizer range conflicts with data bounds |

The results confirm that semantic similarity alone is insufficient, and that integrating data compatibility and execution patterns is critical for practical reliability.

### 4.4.4 RQ4: Learning from execution history

Comparing the full system to *Ablation-NoHistory* quantifies execution history contribution at 5.4 pp ($p=0.007$). Figure 5 presents learning curves from temporal evaluation, showing success rate evolution as execution history accumulates.

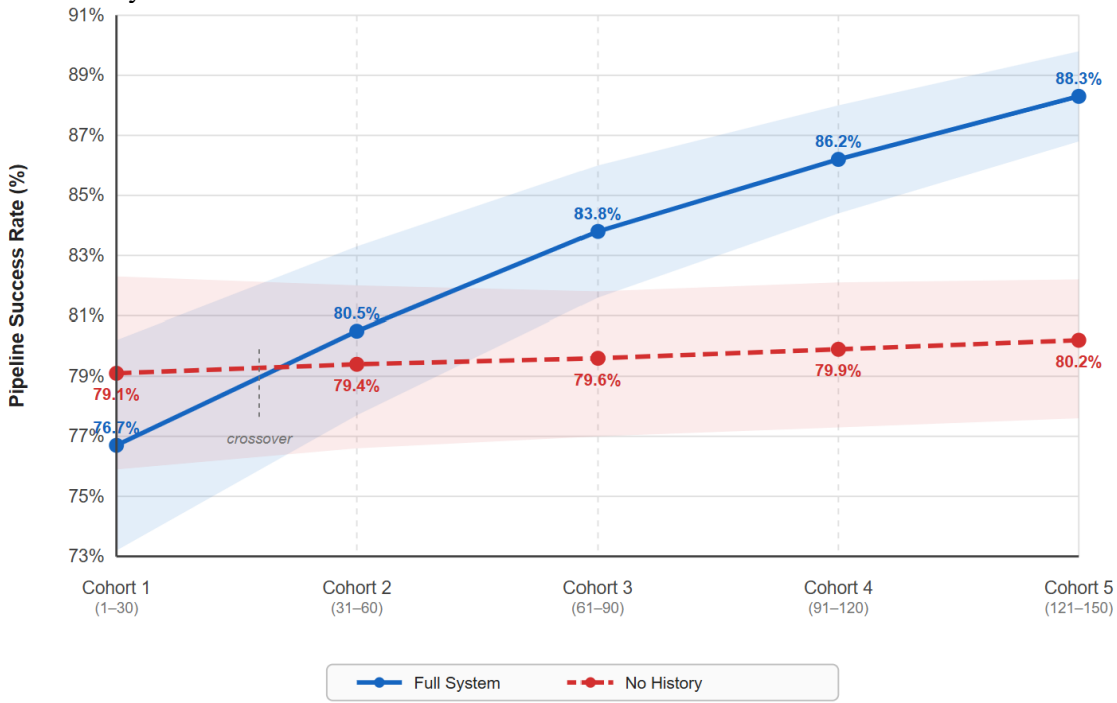

Figure 5. Pipeline success rate evolution across temporal cohorts comparing Full System with execution history learning against No-History ablation

Success rate improves from 76.7% (cohort 1, minimal history, relying on code analysis and compatibility alone) to 88.3% (cohort 5, 120 tasks of accumulated evidence). The NoHistory baseline remains nearly flat (79.1% → 80.2%), confirming that observed improvement stems from learning rather than easier tasks in later cohorts. The 11.6 pp improvement over time demonstrates substantial learning capacity—the system progressively identifies which microservice combinations work well together, which fail frequently, and which workflow patterns succeed for specific task types.

### 4.4.5 RQ5: Self-healing

We injected failures into 20% of executions (30 tasks). LLM-based self-healing recovers 73.3% of failures compared to 23.3% for retry-only ($p<0.001$), with a median added latency of 38 seconds (26.6% overhead relative to typical execution time). Self-healing reasons about root causes, selecting compatible alternatives for type mismatches and less demanding components for configuration errors, while retry-only succeeds only for transient failures. The net effect raises pipeline success on the failure-injected subset from 76.0% to 84.7%, confirming that intelligent alternative selection transforms component failures into manageable substitution events.

## 5. Conclusions

This paper presented an end-to-end autonomous data science framework that transforms a dataset and a NL goal into an executable ML pipeline through five coordinated agents. The experimental evaluation on 150 ML tasks provides evidence addressing each RQ. Regarding RQ1, the system achieves an 84.7% end-to-end pipeline success rate without user intervention in component selection, configuration or error handling, while delivering a 22.7× speedup over manual construction (median 143s vs. 3,240s). Output quality remains competitive with expert-built pipelines (AUC 0.81 vs. 0.84), demonstrating that full automation does not require sacrificing result quality.

For RQ2, code-grounded analysis improves component selection accuracy by 16.6 percentage points over documentation-based discovery, with the largest gains in feature engineering (+19.4 pp), where documentation drift is most prevalent. This validates the architectural decision to treat source code as the authoritative signal for component capabilities, making the system robust to the documentation quality issues that are well-documented in software engineering research.

RQ3 confirms that hybrid scoring significantly outperforms semantic-only ranking (+13.4 pp pipeline success, $p<0.001$). Failure analysis shows that integrating data compatibility and execution patterns prevents 76% of type mismatches and 82% of missing data errors that semantic similarity alone cannot anticipate.

RQ4 demonstrates measurable learning from execution history. Pipeline success rate improves from 76.7% (cohort 1) to 88.3% (cohort 5) as evidence accumulates, while the no-history ablation remains flat (79.1%→80.2%), confirming that improvement stems from learned workflow patterns rather than task ordering effects.

RQ5 shows that LLM-based self-healing recovers 73.3% of injected failures compared to 23.3% for retry-only, with acceptable latency overhead. This mechanism transforms the recommender's ranked candidate list into a recovery queue, enabling graceful degradation rather than pipeline termination.

These results demonstrate that addressing semantic, selection and execution uncertainty within a unified architecture produces a system that exceeds the sum of its individual components. The code-grounded analysis provides reliable representations regardless of documentation quality; the hybrid recommender leverages multiple evidence signals for robust selection; the execution-based learning creates a self-reinforcing feedback loop that continuously improves performance, whereas the self-healing mechanism ensures robustness at runtime.

The remaining 15.3% failure rate arises primarily from ambiguous NL goals, gaps in microservice catalog coverage and runtime incompatibilities not detectable through static analysis, limitations that point toward future work in interactive goal disambiguation, automated microservice generation and dynamic compatibility verification.

**Acknowledgement**. This work was supported by a grant of the Ministry of Research, Innovation and Digitization, CNCS/CCCDI - UEFISCDI, project number COFUND-CETP-SMART-LEM-1, within PNCDI IV. This research was funded by CETP, the Clean Energy Transition Partnership under the 2022 CETP joint call for research proposals, co funded by the European Commission (GAN° 101069750) and with the funding organizations detailed on https://cetpartnership.eu/funding-agencies-and-call-modules.
**Disclosure statement (Competing interest)**. The authors have no relevant financial or non-financial interests to disclose.
**Data availability statement**. Data will be provided based on reasonable request.
**Ethical approval**. Not applicable.
**Informed consent**. Not applicable.
**Author contributions**. **Gabriela Dobrița**: Conceptualization, Methodology, Investigation, Resources, Data Curation, Writing-Original Draft, Validation, Formal analysis. **Adela Bâra**: Conceptualization, Methodology, Formal analysis, Investigation, Resources, Data Curation, Writing-Original Draft, Writing-Review and Editing, Supervision; **Simona-Vasilica Oprea**: Conceptualization, Validation, Formal analysis, Investigation, Method, Writing-Original Draft, Writing-Review and Editing, Visualization, Project administration, Supervision;